\documentclass[runningheads]{llncs}
\usepackage{graphicx}
\usepackage{comment}
\usepackage{amsmath,amssymb} 
\usepackage{color}
\usepackage{multirow}

\usepackage[table]{xcolor}
\usepackage{colortbl}
\def\Gray#1{\cellcolor[gray]{#1}}
\usepackage{sidecap}
\usepackage{caption}

\begin{document}
\pagestyle{headings}
\mainmatter
\def\ECCVSubNumber{3588}  

\title{Learning Propagation Rules \\ for Attribution Map Generation} 

\titlerunning{Learning Propagation Rules for Attribution Map Generation}

\author{Yiding Yang\inst{1}
\and
Jiayan Qiu\inst{2}
\and
Mingli Song\inst{3}
\and
Dacheng Tao\inst{2}
\and
Xinchao Wang\inst{1}
}
\authorrunning{Y. Yang et al.}
\institute{Stevens Institute of Technology, Hoboken, NJ 07030, USA\\
\email{\{yyang99,xinchao.wang\}@stevens.edu}
\and
UBTECH Sydney AI Centre, School of Computer Science, Faculty of Engineering, The University of Sydney, Darlington, NSW 2008, Australia\\
\email{\{jqiu3225@uni.sydney.edu.au,dacheng.tao@sydney.edu.au\}}
\and
College of Computer Science and Technology, Zhejiang University, Hangzhou, China\\
\email{brooksong@zju.edu.cn}}

\maketitle

\begingroup
\setlength{\tabcolsep}{2pt} 
\renewcommand{\arraystretch}{1} 
\begin{figure}[b]
    \centering
    \scalebox{0.72}{
    \begin{tabular}{c c c c c c c c}
    
    \includegraphics[width=19mm]{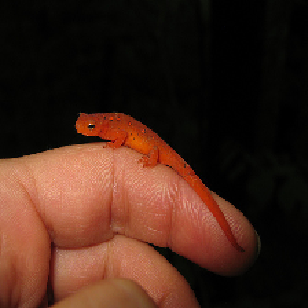} &
    \includegraphics[width=19mm]{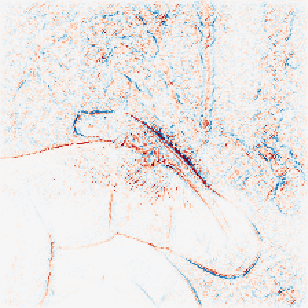} &
    \includegraphics[width=19mm]{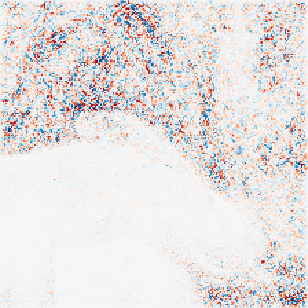} &
    \includegraphics[width=19mm]{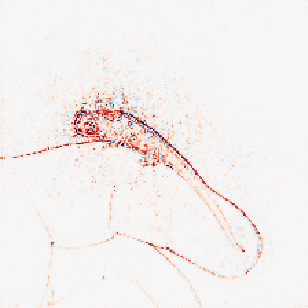} &
    \includegraphics[width=19mm]{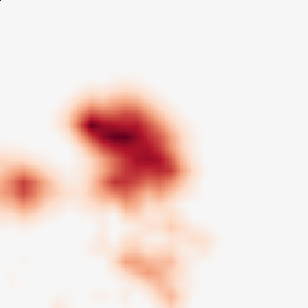} &
    \includegraphics[width=19mm]{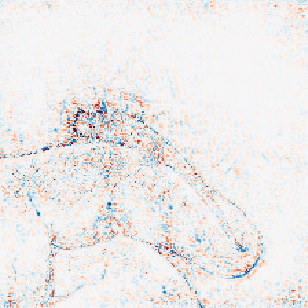} &
    \includegraphics[width=19mm]{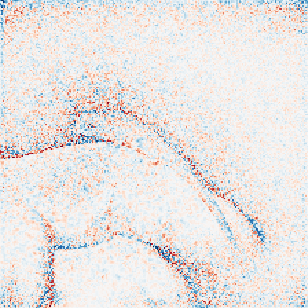} &        \includegraphics[width=19mm]{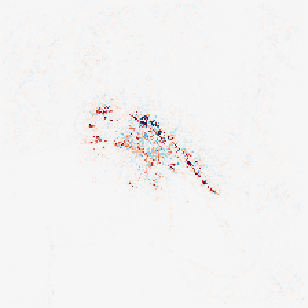}   \\  
    
    Input & 
    DeepLIFT~\cite{shrikumar2017lift} & 
    $\epsilon$-LRP~\cite{binder2016LRP} & 
    GuidedBP~\cite{springenberg2014guidedback_striving} &
    Mask~\cite{fong2017mask} & 
    IG~\cite{sundararajan2017axiomatic} & 
    SG~\cite{smilkov2017smoothgrad}& 
    Ours  \\
    \end{tabular}
    }
    \caption{ What makes this image a \emph{newt}? This figure shows the attribution maps generated by different methods. Existing gradient-based methods fail even in this {simple} case. For example, 
    IG and GuidedBP focus on non-relevant regions, such as the boundary of the hand. Our method, on the other hand, produces cleaner and more focused attribution map.}
    \label{fig:intro}
\end{figure}
\endgroup

\begin{abstract}
Prior gradient-based attribution-map methods rely on hand-crafted 
propagation rules for the non-linear/activation layers
during the backward pass, so as to produce gradients
of the input and then the attribution map.
Despite the promising results achieved, such methods 
{are sensitive to the non-informative high-frequency components 
and lack adaptability for various models and samples.}
In this paper, we propose a dedicated 
method to generate attribution maps that
allow us to learn the propagation rules automatically,
overcoming the flaws of the hand-crafted ones.
Specifically, we introduce a 
learnable plugin module,
which enables adaptive 
propagation rules for each pixel,
to the non-linear layers during the backward pass
for mask generating. 
The masked input image is then fed into the model again to 
obtain new output that can be used as a guidance
when combined with the original one.
The introduced learnable module
can be trained under 
any auto-grad framework with higher-order differential support.
As demonstrated on five datasets and six network architectures, 
the proposed method yields  state-of-the-art results and gives 
cleaner and more 
{visually plausible}
attribution maps.
\keywords{Propagation Rules, Attributions Maps, Learnable Module}
\end{abstract}

\section{Introduction}

Deep learning has made encouraging progress 
and yielded state-of-the-art performances 
in almost all vision and language tasks. 
The gratifying results, however, come
at cost of huge amount of training effort
as well as the often uninterpretable behaviors,
making deep networks {less dependable under some circumstances such as medical image processing}. 
Recently, interpreting deep networks  
has aroused more and more attention from researchers~\cite{adebayo2018sanity,ancona2019polytime_shapleyvalue,ribeiro2018anchors,zhang2018interpreting_aaai1,ghorbani2019interpretation_aaai2,alvarez2018towards_nips1,Feng_2018_NeurIPS,Yu_2017_CVPR,chen2019data,wang2018adversarial,Yang_2020_CVPR,Sheng_2019_AAAI,Ye_2019_CVPR,Ye_2020_CVPR}.
Among the many endeavors,
estimating the \textit{attribution map} has become a mainstream direction.  
The main goal of producing an attribution map is to
{generate a mapping between the pixels and their
corresponding contributions to the prediction, so that the
supports of the prediction can be discovered.}
The work of~\cite{Song_2019_NeurIPS,Song_2020_CVPR}
have also demonstrated that 
attention maps can be utilized
in estimating task transferability.

Existing attribution-map generation methods can be divided into three categories: optimization-based,
perturbation-based, and gradient-based methods. 
Optimization-based methods produce attribution maps
using conventional optimization methods like signal estimation~\cite{kindermans2017patternnet},
and local function approximation~\cite{ribeiro2016shouldlime}.
Such optimizers, however, often require a large number of samples,
making them data-dependent and time-consuming.
Perturbation-based methods, on the other hand, 
produce attribution maps by modifying the input image according
to a mask and then recording the change of output. 
However, they ignore the original gradients of the input.
Gradient-based methods explicitly utilize
gradients of the input for attribution map generation,
and therefore encode the interaction across different pixels,
yielding more informative attribution maps~\cite{montavon2017deeptaylor}.
Given a trained model with fixed parameters, gradients of the input are obtained through loss back-propagation, 
where existing methods focus on designing hand-crafted propagation rules for the non-linear/activation layers~\cite{binder2016LRP,ancona2017unified}.
However, such pre-defined and thus fixed rules
{lack {adaptability} for various models and samples}.

\begin{figure}[t!]
\centering
\includegraphics[width=.98\textwidth]{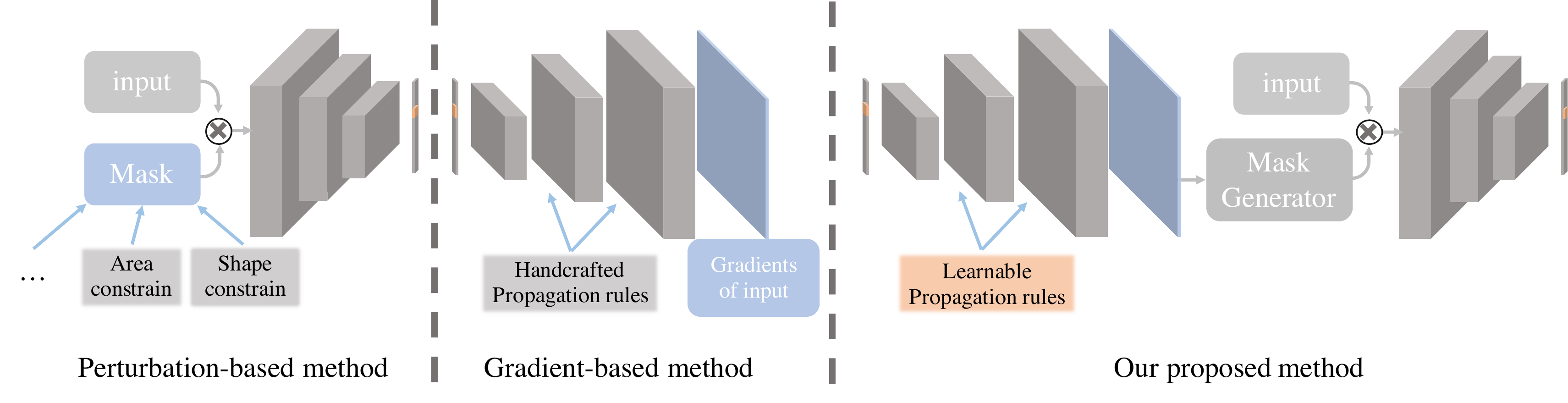} 
\caption{Comparing the perturbation-based method, graident-based method, and our proposed method for generating attribution maps. Different from the perturbation-based method that introduces various constraints to the mask, we generate the mask by making use of the gradients of the input;
Moreover, unlike the gradient-based method that handcrafts the propagation rules, we make them learnable.}
\label{fig:thedifference}
\end{figure}

In this paper, we propose a new method to generate 
the attribution map that makes the propagation 
rules for the non-linear layers \emph{learnable},
and optimize the rules using supervision 
from the model and the input image themselves.
In Fig.~\ref{fig:intro},
we compare our produced attribution map with those
obtained from the state-of-the-art methods. 
Conventional gradient-based methods such as 
DeepLIFT and $\epsilon$-LRP are prone to 
noisy attributions even for the uniformed-colored background.
{Most of the methods focus on non-relevant 
high-frequency regions, such as the boundary of the hand.
Our method, thanks to the more flexible rules, generates a neat and more focused attribution map.}
{Fig.~\ref{fig:thedifference} illustrates
a comparison
of different methods.
Unlike conventional gradient-based approaches that rely on a single 
unifying hand-crafted propagation rule for all models and samples, 
we now make the rules adaptive for any given sample and model.}
Specifically, within our method, each output feature
of the non-linear layers is allowed to behave differently  
during the backward pass, making it possible
to learn more flexible and {advisable} propagation rules
for the attribution map.
As a result, the non-informative regions, 
e.g. the high-frequency ones, 
are suppressed under the supervision,  
leading to a cleaner attribution map.

\begin{figure*}[t!]
\centering
\includegraphics[width=.96\textwidth]{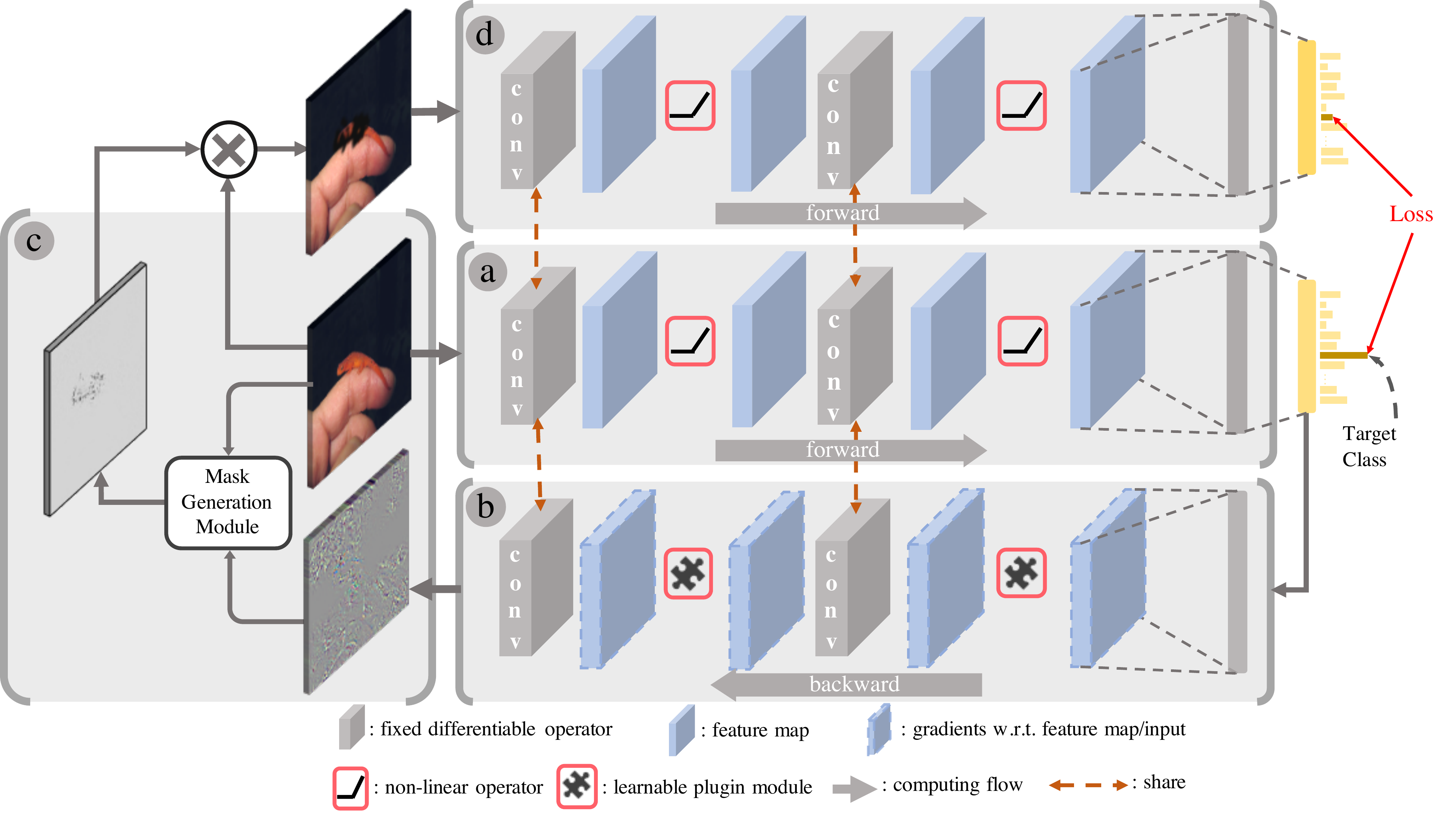} 
\caption{Illustration of our proposed optimization scheme. 
Step $a$ forwards the input image through the trained model to obtain the original activation and its loss. Specifically, we use the pre-softmax/pre-sigmoid output
as the activation of the model.
Step $b$ then backwards the gradients of input image with respect to the prediction loss, in which we introduce the \emph{Learnable Plugin Module}. The attribution map is obtained by the element-wise multiplication between the input image and its gradients. 
Next, step $c$ generates the masks with the attribution map. 
Finally, step $d$ masks the input image and then forwards the masked image through the trained model to get a new activation. {The difference between the new activation and the original one of the target class} serves as the loss, 
which is used to optimize the {learnable plugin module}.}
\label{fig:wholepipeline}
\end{figure*}

To learn the propagation rules,
a new optimization scheme
is proposed as shown in Fig.~\ref{fig:wholepipeline}.
The input image is first fed into a trained
neural network with fixed parameters to obtain the original prediction. 
Then, gradients of the input can be obtained during the
backward pass, in which process, a 
\emph{learnable plugin module}, e.g. neural network, is introduced to control the propagation rules of the non-linear layers.
After obtained the gradients of the input, 
we compute the attribution map and then generate masks for the input.
The input image will then be masked and fed into the trained network for deriving the difference with respect to the original prediction.
Such difference is adopted as a supervision to optimize the learnable plugin module through a new backward pass. 
Since the computation of second-order gradients is required, an auto-grad 
framework with higher-order differential support is used to implement the proposed optimization scheme.

Our contribution is therefore, to our best knowledge,
the first dedicated approach that enables the learning of 
the propagation rules for the non-linear layers
to generate attribution maps.
Unlike the hand-crafted rules,
our method makes it possible to find adaptive propagation 
rules for any given model and sample. 
The learning of the rule is achieved via a novel optimization scheme:
{the learnable module we introduced 
can be optimized under any auto-grad framework 
with higher-order differential support.}
We conduct experiments on three different datasets and six models with different architectures.
Our proposed method yields state-of-the-art results 
and produces a cleaner attribution map.

\section{Related Work}

Here we give a brief description of the related work. We start by reviewing the  attribution-map methods of three categories: optimization-based, gradient-based, and perturbation-based methods. 
We then discuss the higher-order differential algorithms for implementing our proposed method. Note that the proposed method differs from all three categories of attribution-map methods. 
Specifically, 
compared with optimization-based methods, our proposed method depends only on  given samples; 
compared with gradient-based methods, our proposed method enables the learning of the propagation rules; 
compared with perturbation-based methods, 
our proposed method involves the gradients of inputs 
{as the condition to generate the mask rather than the human designed constraints}. 

\textbf{Optimization-based methods. }
These methods adopt conventional optimization scheme
to generate an attribution map.
For example, PatternNet~\cite{kindermans2017patternnet} designs a signal detector to filter out the non-informative components. Then, a quality measurement criterion is introduced to optimize the attribution map generation.
Instance feature selector~\cite{chen2018learning} learns a feature selector by
maximizing the mutual information between the selected features and the model's
response.
Another method, LIME \cite{ribeiro2016shouldlime}, locally approximates a non-linear model with a linear function on the given sample,
and then generates the attribution map from the linear function.
However, these methods are data-dependent and time-consuming.

\textbf{Gradient-based methods.}
Such methods utilize gradients of 
the input to generate an attribution map.
For example, Deep saliency~\cite{simonyan2013saliencydeep} 
generates the attribution map by backwarding the loss with respect to the input and taking the
absolute value of the gradients.
Moreover, the element-wise multiplication between the input and its gradients improves the performance~\cite{shrikumar2016gradientxinput}.
Another method, Guided backpropagation~\cite{springenberg2014guidedback_striving}, 
shows that ignoring the negative gradients helps to distinguish the contribution of each pixel.
As for DeepLIFT~\cite{shrikumar2017lift}, reference features are added to the non-linear layers to reduce the influence from baseline. 
However, the fixed propagation rules of these methods lack adaptability for various models and samples.

\textbf{Perturbation-based methods.}
Methods along this line make the assumption that removing important pixels will degrade the prediction accuracy,
{and generate the mask based on some constraints.}
One of the methods, Occlusion~\cite{zeiler2014occlusion_visualizing},
generates the attribution map by systematically occluding different parts of the input image and then recording the change of output. 
Moreover, it is also possible to obtain an occlusion mask by learning~\cite{fong2017mask} rather than brute force searching. sMask~\cite{fong2019understanding} introduce predefined area constrain and smooth constrain.

\textbf{Higher-order differential algorithms.}
High-order differential algorithms make the second-order
gradients computation possible~\cite{griewank2008evaluatingdifferential01,maclaurin2016modelingdifferential02}, and are thus essential for our proposed attribution-map method. 
Most of the current deep learning libraries implement these algorithms.
In Pytorch, for example, gradients of a variable remains a variable,
which enables the computation of higher-order gradients 
by recursively computing the first-order gradients. 
These implementations serve as
the backbone of our proposed optimization scheme.

\section{Method}
Propagation rules for the non-linear layers
during the backward pass can be treated as a pixel-to-pixel
mapping between gradients of the adjacent feature maps. 
For gradient-based {attribution-map} methods, they are proven to be unified and distinguished only on the propagation rules of the non-linear layers~\cite{ancona2017unified}.
However, existing gradient-based methods formulate the gradient computation as hand-crafted rules,
which are hard to fit all models and samples. For example, a method may work well
for images with clean backgrounds but fail on those with complex backgrounds. 

To this end, we propose a novel method, which makes the rules learnable.
The rules will be optimized {individually} under the supervision
of every combination of the input image and the trained neural network,
making it adaptive for  given models and samples.
Fig.~\ref{fig:wholepipeline} illustrates the
proposed optimization scheme, which is also based on the gradient-descent optimization method.
There are four steps in a single optimization iteration 
for attribution map generation. 
{
In Step~1, we forward the input image throughout the network;
in Step~2, we backward gradients of the input through our learnable plugin module for attribution map generation;
in Step~3, we generate the mask with the attribution map;
in Step~4, we forward the masked input image to obtain the loss
by computing the difference of activations between the original input and
the masked one, where the activation is referred as the 
pre-softmax/pre-sigmoid output of the model.}

\subsection{Step~1: Forwarding the input image} 

In this step, the input image is fed into 
a trained neural network model to obtain the activation. 
Once the target class to be interpreted is chosen,
we set the gradient of the activation to one for target class
and zero for the others which is commonly used in
gradient-based methods~\cite{springenberg2014guidedback_striving,shrikumar2016gradientxinput,binder2016LRP}.
We then pass the gradient to the next step.

\subsection{Step~2: Backwarding the gradients of the input}

Given the gradient of the model's activation from previous step, the gradients of the input image can be computed through the back-propagation algorithm. During this process, we implement our proposed {learnable plugin module} to learn the propagation rules for non-linear layers, instead of modifying the hand-crafted function.

Specifically, after feeding a feature map $f_{in}$ into a non-linear function $g$, 
we will have the output $f_{out}=g(f_{in})$. 
Let $\mathcal{L}$ denotes the training loss during the backward pass.
The gradient of $f_{out}$ is the partial derivatives 
of $\mathcal{L}$ with respect to the output feature map 
that be expressed as
$\frac{\partial \mathcal{L}}{\partial f_{out}}$.

Then, following the chain rule of the back-propagation algorithm, the partial derivatives of $\mathcal{L}$ with respect to $f_{out}$, can be computed as
\begin{equation}
    \frac{\partial \mathcal{L}}{\partial f_{in}} = 
   \frac{\partial \mathcal{L}}{\partial f_{out}} \cdot \frac{\partial g(x)}{\partial f_{in}} = 
   \frac{\partial \mathcal{L}}{\partial f_{out}} 
   \cdot \frac{\partial g(f_{in})}{\partial f_{in}},
\end{equation}
where $\frac{\partial g(x)}{\partial x}$ denotes the derivation of non-linear function $g$ with respect to it's input.

In existing gradient-based methods, $\frac{\partial g(x)}{\partial x}$ is
manually modified for different purposes,
such as ignoring the neurons that suppress the target output~\cite{springenberg2014guidedback_striving}. 
Although numerous hand-crafted propagation rules are proposed,
none of them is optimal for all scenarios.
For example, some hand-crafted rules 
are unresponsive to certain target class,
while others may be too sensitive to 
ignore non-relevant high-frequency components.

Therefore, instead of using a fixed hand-crafted $\frac{\partial g(x)}{\partial x}$, 
we introduce a learnable plugin module, denoted as $\mathit{G}$,
as the basic modules.
This module takes the gradients of feature map $\frac{\partial \mathcal{L}}{\partial f_{out}}$ from the upper layer
and computes $\frac{\partial \mathcal{L}}{\partial f_{in}}$ as
\begin{equation}
    \frac{\partial \mathcal{L}}{\partial f_{in}} = \mathit{G}( \frac{\partial \mathcal{L}}{\partial f_{out}} )
\end{equation}
where $G$ can be plug-and-play
without modifying the original architecture
of the given trained neural network.
{We provide two architectures for $G$, 
For the first architecture, we have $C$ parameters per layer, 
where $C$ is the number of channels. 
{The operation of $G$}
is similar to a standard convolutional operation
without the sum operation, which shares parameters across 
different positions within a same layer.
For the second architecture, we do not share parameters across different positions,
leading to more flexible control of the rules at cost of
more parameters.}
Once the gradients of the input are obtained, 
the attribution map can be computed as
\begin{equation}
    {\mathcal{A}} = \frac{\partial \mathcal{L}}{\partial I} \circ I,
\end{equation}
where ${\mathcal{A}}$, $I$, and $\frac{\partial \mathcal{L}}{\partial {I}}$ denote the attribution map, 
input image, its gradients respectively, 
and $\circ$ is Hadamard product. 
The pros and cons of multiplying the generated gradient with the input
have been discussed in ~\cite{smilkov2017smoothgrad}.
In this paper, this multiplication is adopted across all experiments.

\subsection{Step~3: Mask generation}

Given the attribution map, a mask can be generated to 
segment out the image parts that contribute most to the target-class recognition. Since the distribution of attribution map varies a lot, we propose a \emph{Mask Generation Module} for generating suitable masks here.

\begin{SCfigure}
\centering
\includegraphics[width=.43\textwidth]{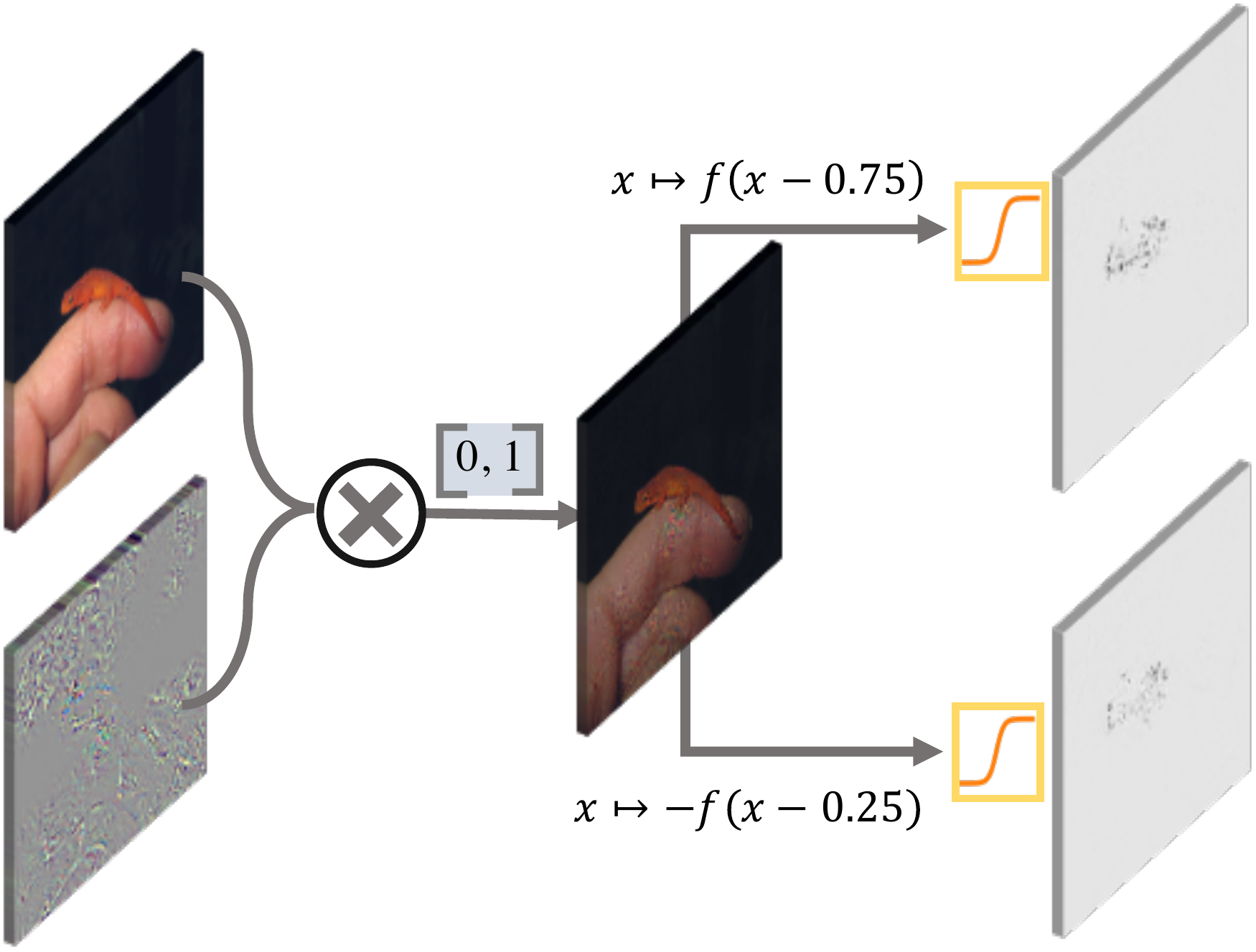} 
\caption{Illustration of the mask generation module. First, 
the attribution map is generated by element-wise multiplication 
between the input image and its gradients.
Then, the two masks are computed by feeding the shifted 
and scaled attribution map into a sigmoid function.}
\label{fig:mask_gen}
\end{SCfigure}

It can be seen from Fig.~\ref{fig:mask_gen} that 
the attribution map is first scaled to $[0,1]$ and then shifted to a fixed center. Finally, we implement the mask generation using a sigmoid function. We write,
\begin{equation}
    \mathcal{M}^p = 1 - \frac{1}{1+e^{(- \gamma * ({\mathcal{A}}-\alpha) )} },
\end{equation}
\begin{equation}
    \mathcal{M}^n = \frac{1}{1+e^{(-\gamma * ({\mathcal{A}}-\beta) ) } },
\end{equation}
where $\mathcal{M}^{p}$ denotes the positive mask that segments
pixels with positive contribution, $\mathcal{M}^n$ denotes the
negative mask that segments pixels with negative contribution, 
$\alpha$, $\beta$ denote the fixed centers and 
$\gamma$ denotes the scale factor for sharpening the mask. 

Notice that many other mask generation strategies can be 
directly adopted here, leading to different properties of the 
generated attribution map. For example, in order to generate
smoother mask, a Gaussian smooth function can be replaced
for the previous mask generation functions, which can be written as:

\begin{equation}
    \mathcal{M}^p = \frac{\sum_{v\in \mathcal{A}}\mathcal{S}_{\sigma}(u-v) \mathcal{A}(v)}
                    { \sum_{v\in \mathcal{A}} \mathcal{S}_{\sigma}(u-v) },
                    \mathcal{S}_{\sigma}(u) = e^{-\frac{\|u\|^2}{2\sigma^2}}
\end{equation}
where $\mathcal{A}$ is the generation attribution map, $u, v$ is the index of values in $\mathcal{A}$
and $\mathcal{S}$ is the smooth function.

\subsection{Step~4: Forwarding the masked image}
{Based on the generated masks, pixels with special contributions 
will be segmented out from the input image to form the masked image.}
We then forward the masked image through the trained model to obtain the difference between the activation of input image and masked image, which is used as the loss to train the {learnable plugin module}.

Some pixels in the masked image contribute to correct prediction
while some pixels degrade the accuracy. 
Therefore, instead of measuring the contribution according to the sign of gradients, we propose a \emph{sign-aware loss} to distinguish the pixels.

There are two terms in the sign-aware loss, a positive term and a negative one. The positive term $\mathcal{L}^p$ considers the positively contributed components of the input image:
\begin{equation}
    \mathcal{L}^p = \mathit{F}_t(I \circ \mathcal{M}^p) - \mathit{F}_t(I),
\end{equation}
where $\mathit{F}$ denotes the trained model, $t$ denotes the index of the target class, and $\mathit{F}_t(I)$ denotes the predicted possibility of $I$ to be class $t$. 
By the same token, the negative term is defined as
\begin{equation}
    \mathcal{L}^n = \mathit{F}_t(I) - \mathit{F}_t(I \circ \mathcal{M}^n).
\label{eq:ln}
\end{equation}
This term is defined as the difference of predictions between the input image and the negatively masked image, because deleting the negatively contributed pixels should improve the prediction accuracy. 

{To avoid the trivial solutions like an all-zero mask or an all-one mask}, we introduce mask loss to constrain the strength of the generated masks:
\begin{equation}
    \mathcal{L}^m = |\mathbf{1} - \mathcal{M}^p| + |\mathbf{1} - \mathcal{M}^n|,
\end{equation}
where $|\bullet |$ denotes the $L_{1}$-Norm, $\mathbf{1}$ denotes matrix with all ones.  
We thus have the final loss function:
\begin{equation}
    \mathcal{L} = (\mathcal{L}^p + \mathcal{L}^n) + \lambda \cdot \mathcal{L}^m,
\end{equation}
where $\lambda$ is the hyper-parameter for loss balancing.

\section{Experiments}

\subsection{Evaluation protocols}
There are two kinds of attribution-map errors: error from the attribution-map method itself and error from the trained model.
We conduct objective evaluation which focuses purely on the first type of error.
We adopt most important relevant features~(MoRF) curve as one of the metrics, 
as done by many other methods~\cite{samek2016rf_evaluating,bach2015LRP,binder2016LRP}.
Specifically, we first sort the pixels in an ascending manner 
according to the attribution map,
and then obtain the MoRF curve by incrementally computing the 
correlation between the {different} ratios of
the activation and the ratios of masked pixels.
We also adopt least important relevant features~(LeRF) curve as one objective evaluation metric. LeRF is of the same setting as MoRF except it first sorts the pixels in a descending manner.
For all objective evaluations, we set the upper limit of the number of masked pixels 
to be 5\% of the entire input image. 
We derive the MoRF curve by averaging 1,000 random samples for stability.

ROAR~\cite{hooker2019benchmark} is another metric to evaluate the performance of attribution map.
ROAR first replaces fraction of the pixels that are estimated by the attribution map as the most important
ones with uninformative value. Then, the modified data are used to retrain the same model from scratch
and test it on the modified test set. It claims that a good attribution map should lead to a sharp degradation
of the performance on the modified dataset.

\subsection{Implementation Details}
\paragraph{\bf Attribution-map framework.}
We build a PyTorch-based attribution-map toolbox, which implements the proposed optimization scheme and some of the compared methods. In our toolbox, the gradient-based methods are unified by sharing the forward and backward hook functions. In the objective experiments, since our only concern is about the model interpretation, the class with the highest prediction possibility is set as the ground truth. We then adopt the negative log likelihood function as the loss function.
The running time of our method for an ImageNet-like input 
with a VGG-16 model is about 3s using an Nvidia 1080Ti GPU.

\paragraph{\bf Learnable plugin module.}
For the objective experiments, we use the second architecture of
\emph{learnable plugin module}. 
Specifically, we set the parameter matrices of 
{learnable plugin module} to be of the same size as the input feature map. 
For every non-linear layers in the model, 
{learnable plugin module} first computes the Hadamard product between its parameter matrix and the input feature map, then follows a tanh activation function, 
We also conduct some experiments of the first archtecture of the learnable plugin module which
is similar as the convolution without summation and shares parameters within one layer.
Although the structure is concise, it leads to a significant improvement of the performance. 
The learnable plugin module is optimized with Adam~\cite{kingma2014adam}.
{Similar to sMask~\cite{fong2019understanding}, we train the plugin module separately
for each sample. The plugin modules within different layers do not share parameters and are placed in every nonlinear layer within two convolution layers.}

\paragraph{\bf Reference baseline.}
In order to improve the flexibility of 
{learnable plugin module}, we adopt the reference baseline from DeepLIFT~\cite{shrikumar2017lift}.
Specifically, an additional all-zero input is added in the forward pass to obtain the reference feature maps. Then, all the original features are modified by subtracting their corresponding reference feature map.
\paragraph{\bf Hyper-parameters settings.}
We adopt same hyper-parameters to all experiments, with $\lambda$ set to be 0.1, $\alpha$ set to be 0.75, $\beta$ set to be 0.25, and $\gamma$ set to be 10. For the Adam optimizer, we set the learning rate to be 0.2. No weight decay is used. 
{The performance with respect to} these hyper-parameters are stable within a large value range, and the analysis will be presented in the sensitivity analysis section.

\subsection{Compared Methods}
Here, we give a brief description of the compared methods.
\begin{itemize}
    \item {\bf Gradient-based methods}. GradientXInput~(GradXIn) \cite{shrikumar2016gradientxinput} and DeepSaliency (Saliency)~\cite{simonyan2013saliencydeep} generate  attribution maps from the gradients of the input. Specifically, DeepSaliency ultilizes the gradients only, while GradXIn uses both the input and its gradients. $\epsilon$-LRP~\cite{bach2015LRP}, and
    DeepLIFT~\cite{shrikumar2017lift},
    on the other hand, focus on designing hand-crafted fixed propagation rules to enhance performances.
    SQ\_SG~\cite{hooker2019benchmark} is an improvement over smooth grad method by averaging the squared gradients.
    Integrated Gradients~(IG)~\cite{sundararajan2017axiomatic}
    and Smooth Gradients~(SG)~\cite{smilkov2017smoothgrad} 
    compute the average gradients of multiple inputs by introducing integration paths and adding noises, respectively.
    
    \item {\bf Perturbation-based methods}. Mask~\cite{fong2017mask} treats the attribution map as a mask and learns it {from} a designed framework. sMask~\cite{fong2019understanding} adds more constraints to the mask. RISE~\cite{petsiuk2018rise} generates random masks and obtains the 
    attribution map by linear combining these masks according to outputs of masked images.
    
\end{itemize}

The Mask method is tested on ImageNet dataset only, because it is designed for ImageNet-similar images. The hyper-parameters of Mask are set according to their published work. For the SmoothGrad method, the number of random {noised} images is set to 50. As for the IntegratedGrad method, {the number of integrated images along
the integral path from zero baseline to the original input is set to 50.}
The tolerance of pointing game is set to 15 for all compared methods.

\subsection{Experimental Results}

In this section, we first {analyze} the comparison results between our proposed method and the compared 
methods in the aspects of MoRF, LeRF, and ROAR. 
Then, we present the case study, the sensitivity analysis of hyper-parameters, and the ablation study to completely evaluate our proposed method.

\begin{figure}[]
	\centering
	\includegraphics[width=1.0\textwidth]{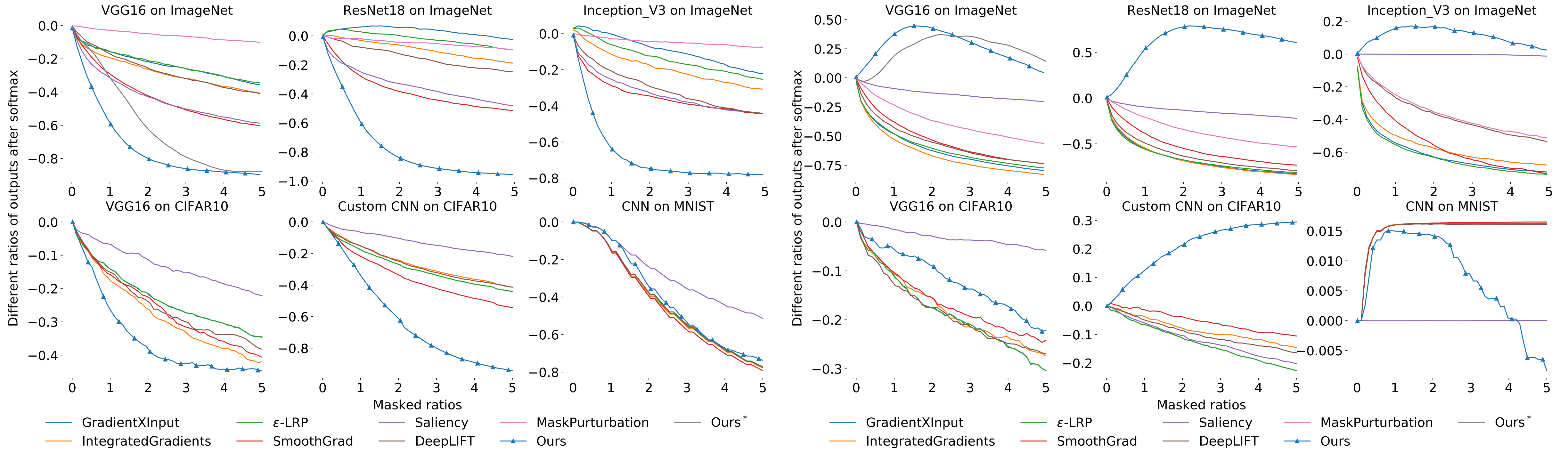}
	\caption{MoRF~(left) and LeRF~(right) curves. The x-axis represents the masked ratio of the entire image and the y-axis represents
		 the difference ratio of the activation after masking the input. 
		Our method produces consistently steeper curves, especially for complex models and samples. Note that for
		the LeRF metric, our method provides the correct negative attributions, which will lead to an increase of the prediction accuracy when removing them. All other methods, as a comparison, fail and still generate positive attributions. }
	\label{fig:most_important}
\end{figure}

\paragraph{\bf MoRF/LeRF result analysis.}

\begingroup
\setlength{\tabcolsep}{2pt} 
\renewcommand{\arraystretch}{1} 
\begin{figure*}[]
	\centering
	\scalebox{0.62}{
		\begin{tabular}{c | c c c c c c c c c}
			 Original 
			&  DeepLIFT~\cite{shrikumar2017lift} 
			&  GradXIn~\cite{shrikumar2016gradientxinput} 
			&  Mask~\cite{fong2017mask} 
			&  IG~\cite{sundararajan2017axiomatic}
			&  Saliency~\cite{simonyan2013saliencydeep} 
			&  SG~\cite{smilkov2017smoothgrad} 
            &  GCAM~\cite{selvaraju2017gradcam}
            &  sMask~\cite{fong2019understanding}
			&  Ours \\
			
			\includegraphics[width=19.4mm]{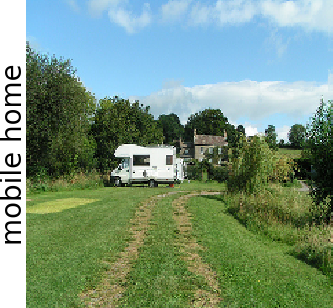} &
			\includegraphics[width=18mm]{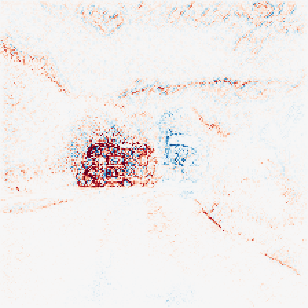} &
			\includegraphics[width=18mm]{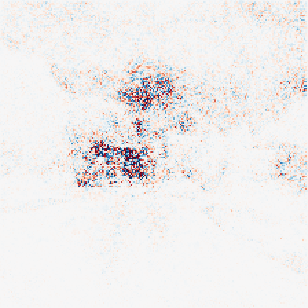} & 
			\includegraphics[width=18mm]{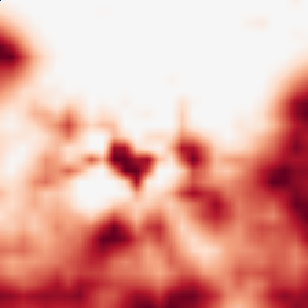} & 
			\includegraphics[width=18mm]{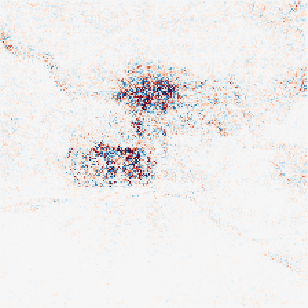} &
			\includegraphics[width=18mm]{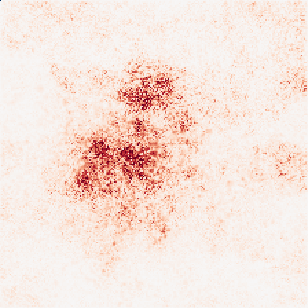} &
			\includegraphics[width=18mm]{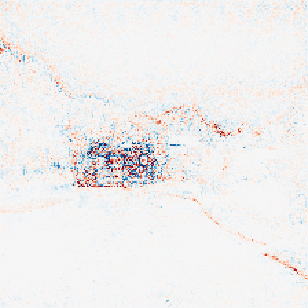} &
			\includegraphics[width=18mm]{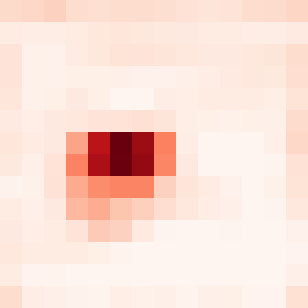} &
			\includegraphics[width=18mm]{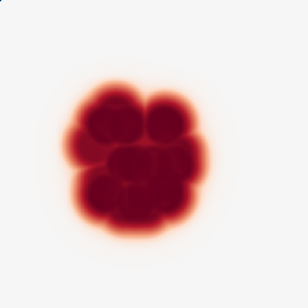} &
			\includegraphics[width=18mm]{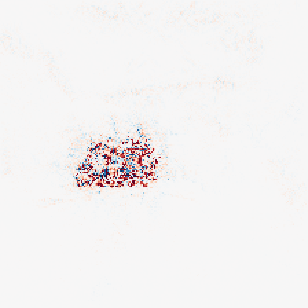} \\
			
			\includegraphics[width=19.4mm]{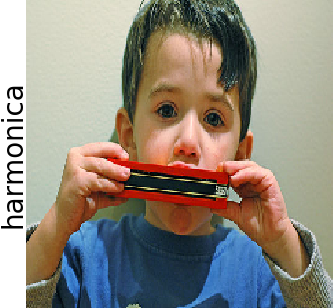} &     
			\includegraphics[width=18mm]{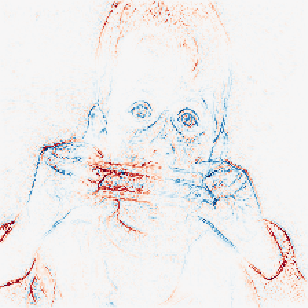} &
			\includegraphics[width=18mm]{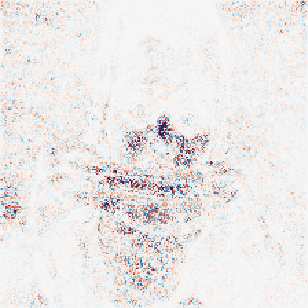} & 
			\includegraphics[width=18mm]{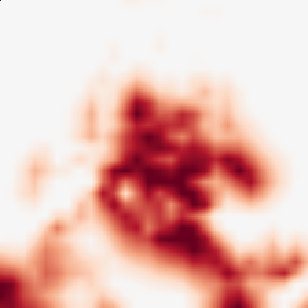} & 
			\includegraphics[width=18mm]{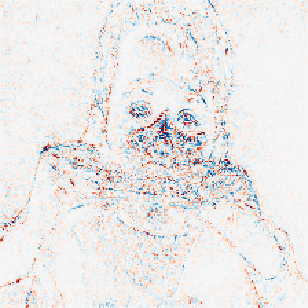} &
			\includegraphics[width=18mm]{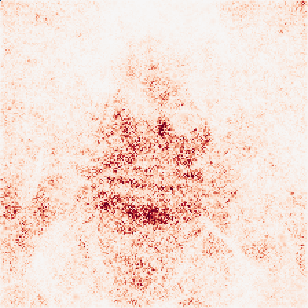} &
			\includegraphics[width=18mm]{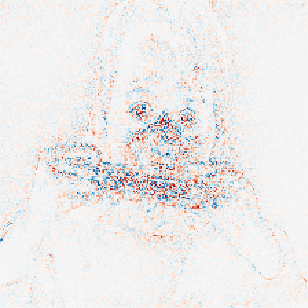} &
			\includegraphics[width=18mm]{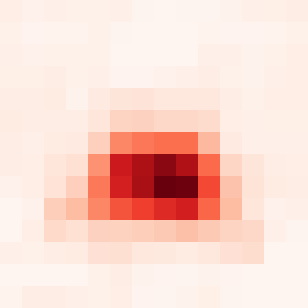} &
			\includegraphics[width=18mm]{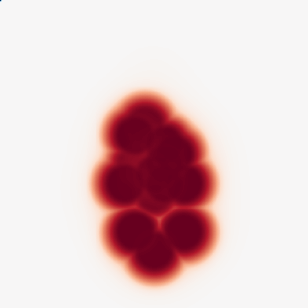} &
			\includegraphics[width=18mm]{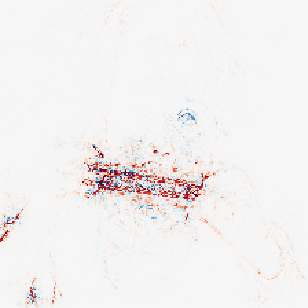} \\
			
			\includegraphics[width=19.4mm]{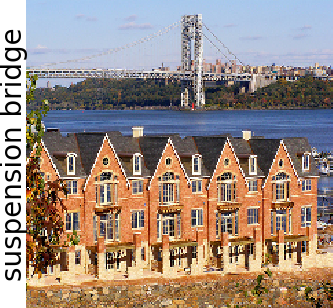} &     
			\includegraphics[width=18mm]{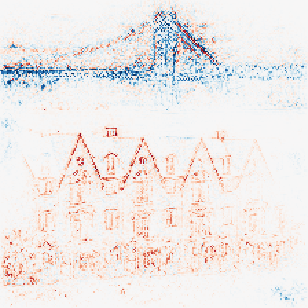} &
			\includegraphics[width=18mm]{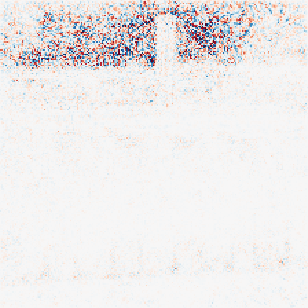} &  
			\includegraphics[width=18mm]{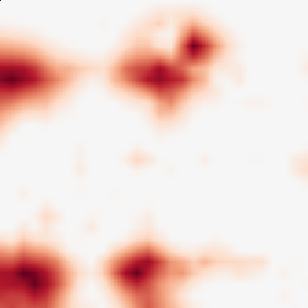} & 
			\includegraphics[width=18mm]{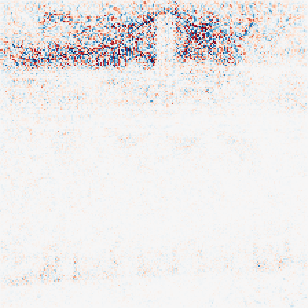} &
			\includegraphics[width=18mm]{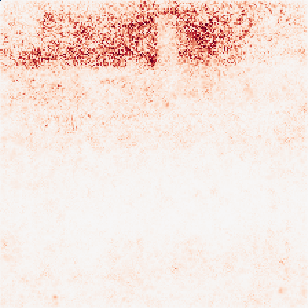} &
			\includegraphics[width=18mm]{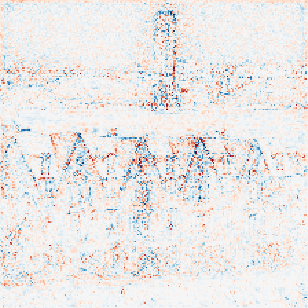} &
			\includegraphics[width=18mm]{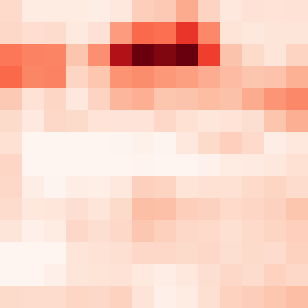} &
			\includegraphics[width=18mm]{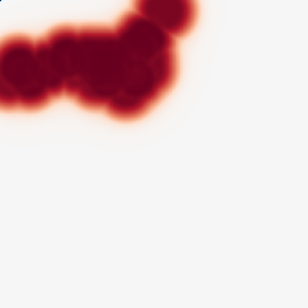} &
			\includegraphics[width=18mm]{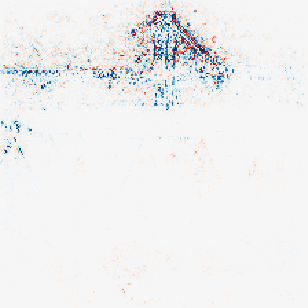} \\

		\end{tabular}
	}
	\caption{Visualization of the attribution maps generated by different methods
		on ImageNet dataset. 
		The attribution maps generated by our method
		are cleaner and {more} visually plausible.
		For gradient-based methods, some of them are {too sensitive to ignore} the non-relevant high-frequency components while others are not responsive enough to the target class~(e.g. the last line). More visualization results will be presented in the supplemental material.
	}
	\label{fig:image_vis}
\end{figure*}
\endgroup

Fig.~\ref{fig:most_important} compares the MoRF and LeRF curves of our proposed method and the compared methods. We conduct the objective experiments on three widely used datasets, MNIST, CIFAR-10, and ImageNet. $*$ means the plugin module with the first architecture which shares parameters across different positions. The analysis of the results on the three datasets is as follows:
\begin{itemize}
    \item {\bf MNIST} is a relative small and simple dataset, 
    in which the images contain only unit digits with clean background. 
    We test on it using a customized CNN model with two convolutional layers. It can be seen that all methods lead to similar performances on both the MoRF and LeRF curves. This can be in part explained by the fact that, simple images cannot distinguish the potentials of these methods.
    
    \item {\bf CIFAR-10} is a larger dataset, in which the images contain 
    common objects but with low resolution. 
    We test two CNN models on this dataset,
    a custom CNN model with four convolutional layers and a VGG-16 model~\cite{simonyan2014vgg}.
    It can be seen that our proposed method performs the best on both the MoRF and LeRF curves. 

    \item {\bf ImageNet} is one of the largest and most complex datasets, in which the images come from the real-world scenes. We test all methods with three state-of-the-art models on this dataset, a VGG-16 model, a ResNet18 model~\cite{he2016resnet}, and a Inception-V3 model~\cite{Szegedy_2016_CVPR_inception_v3}. It can be seen that our method performs the best on the MoRF curve consistently by a large margin. As for the LeRF cure, thanks to the sign-aware loss, our method gives the correct negative attributions while all compared methods fail. 
\end{itemize}

\begin{SCtable}
	\centering
	\scalebox{0.9}{
		\begin{tabular}{c|c c c c}
			Fraction & Original & Random & SQ\_SG & Ours  \\ \hline
			40\% & 80.73\% & 73.65\% & 74.90\% & 71.83\% \\ 
			50\% & 80.73\% & 72.46\% & 72.97\% & 69.81\% \\ 
			60\% & 80.73\% & 70.89\% & 70.79\% & 67.32\% \\ 
			70\% & 80.73\% & 68.98\% & 68.33\% & 63.80\% \\ 
			80\% & 80.73\% & 65.85\% & 65.19\% & 59.08\%  \\ 
			90\% & 80.73\% & 59.58\% & 57.47\% & 50.72\% \\ 
		\end{tabular}
	}
	\caption{Evaluation of ROAR on the CIFAR10 dataset with the custom CNN model. Our method
		consistantly performs better that others especially when the fraction of 
		removed pixels are large. We report the test accuracy on the modifed dataset~(lower is better).}
	\label{tab:roar_res}
\end{SCtable}
\vspace{-1.2em}

\begin{figure}[h]
\centering
\includegraphics[width=1.0\textwidth]{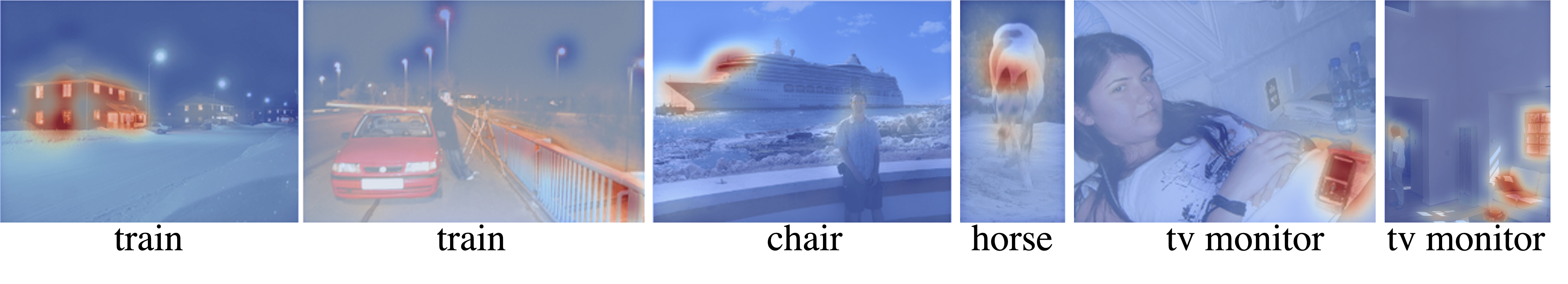} 
\vspace{-1.0em}
\caption{Illustration of some misclassified images and their corresponding attribution maps. 
	The predictions of these images are train, train, chair, horse, tv monitor and tv monitor respectively. 
Red color highlights the supports of the predictions. Note that although the attribution map
does not align well with the human perception, masking the image according to the attribution map still lead to a significant drop of the wrong prediction.}
\label{fig:misvis}
\end{figure}
\vspace{-1.2em}

We visualize the attribution maps on ImageNet dataset with a trained VGG-16 model in Fig.~\ref{fig:image_vis}. It can be seen that some compared methods are too sensitive to ignore the non-relevant high-frequency components (e.g. the grass and the boundary of trees). Other methods, unfortunately,  fail to localize the most contributed areas to the prediction (e.g. the results in the last line). 
As a comparison, our method provides consistently cleaner and more focused attribution maps.

\paragraph{\bf ROAR result analysis.} 
Tab.~\ref{tab:roar_res} shows the ROAR result. We change the fraction of removed pixels from
40\% to 90\% and retrain the model on the modified dataset. The test accuracy is reported and
lower is better. Our method consistantly leads to a lower accuracy on the modified test set.

\paragraph{\bf Case study.}
We first conduct a case study of some miscalssfied images shown in Fig.~\ref{fig:misvis}. It can be seen that even the generated attribution map does not align well with the
human perception, masking the input according to the attribution map
stills lead to a significant drop of the wrong prediction.
We also conduct a case study on a composite image. It contains two objects from different classes. 
Our method is truly responsive to the target class,
as can be seen from Fig.~\ref{fig:composite_vis},
and focuses on the most informative areas while many gradient-based methods are not sensitive to the target class or even give the opposite signs~(like DeepLIFT).

\begingroup
\setlength{\tabcolsep}{2pt} 
\renewcommand{\arraystretch}{1} 
\begin{figure*}[]
\centering
\scalebox{0.66}{
\begin{tabular}{c | c c c c c c c c}
     Original 
    &  DeepLIFT~\cite{shrikumar2017lift} 
    &  GradXIn~\cite{shrikumar2016gradientxinput} 
    &  Mask~\cite{fong2017mask} 
    &  GuidedBP~\cite{springenberg2014guidedback_striving} 
    &  IG~\cite{sundararajan2017axiomatic} 
    &  Saliency~\cite{simonyan2013saliencydeep} 
    &  SG~\cite{smilkov2017smoothgrad} 
    &  Ours \\
    
    \includegraphics[width=18mm]{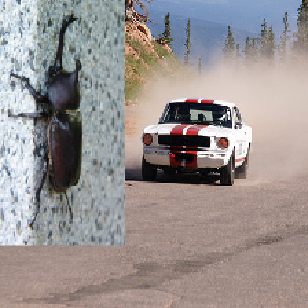} &     
    \includegraphics[width=18mm]{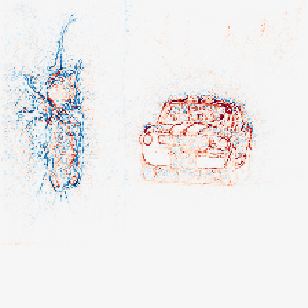} &     
    \includegraphics[width=18mm]{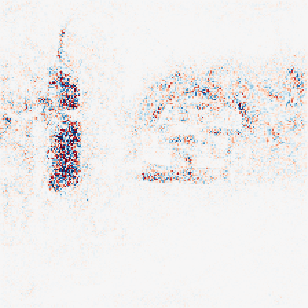} &
    \includegraphics[width=18mm]{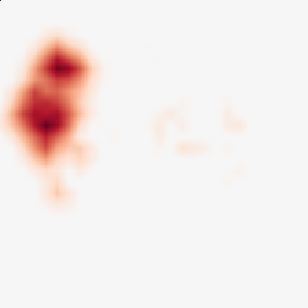} &
    \includegraphics[width=18mm]{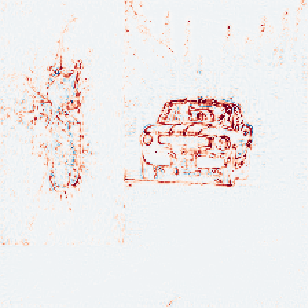} &
    \includegraphics[width=18mm]{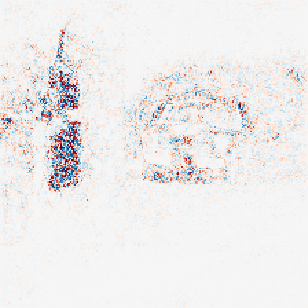} &
    \includegraphics[width=18mm]{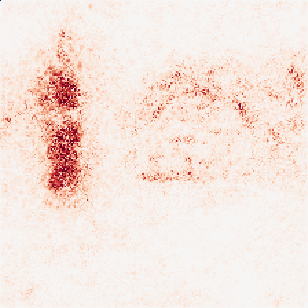} &
    \includegraphics[width=18mm]{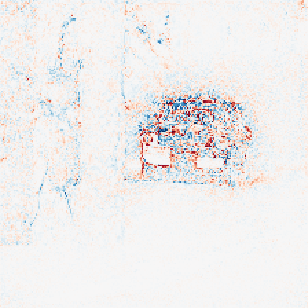} &
    \includegraphics[width=18mm]{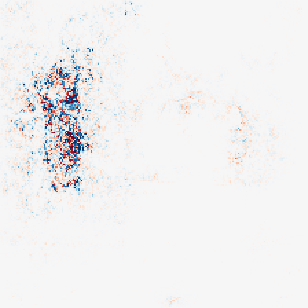} \\
    
     &     
    \includegraphics[width=18mm]{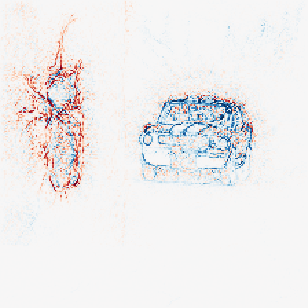} &     
    \includegraphics[width=18mm]{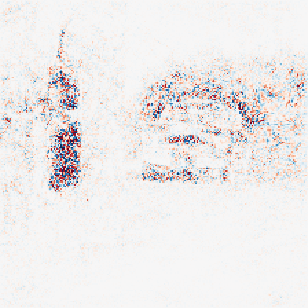} &
    \includegraphics[width=18mm]{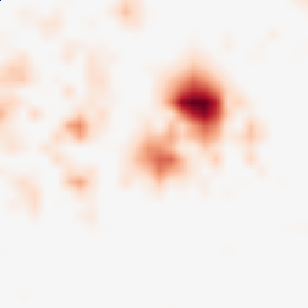} &
    \includegraphics[width=18mm]{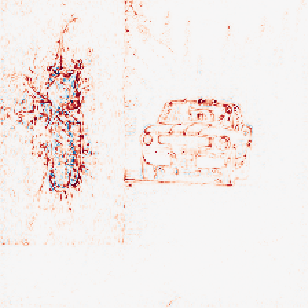} &
    \includegraphics[width=18mm]{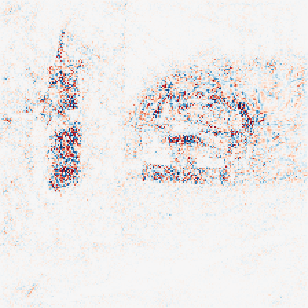} &
    \includegraphics[width=18mm]{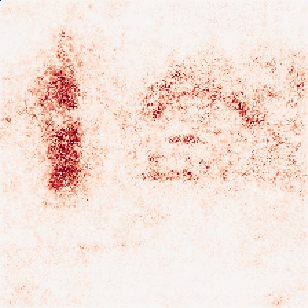} &
    \includegraphics[width=18mm]{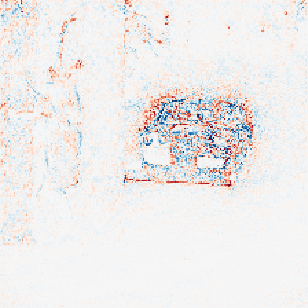} &
    \includegraphics[width=18mm]{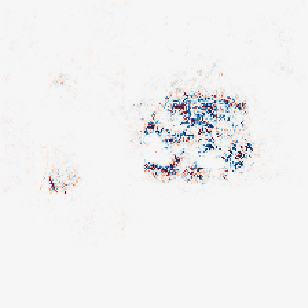} \\
    
\end{tabular}
}
\vspace{-1.0em}
\caption{Visualization of a composite image with two different objects.
The target name of the first line is Rhinoceros beetle while the 
target name of the second line is Landrover. 
Many gradient-based methods are not sensitive to the target class.}
\label{fig:composite_vis}
\end{figure*}
\endgroup
\vspace{-1.3em}

\paragraph{\bf Sensitivity analysis.}
Here, we conduct experiments for analyzing the sensitivity of the four hyper-parameters in our method including the scale factor $\gamma$, two fixed centers $\alpha$ and $\beta$, and $\lambda$ for loss balancing. All results are obtained using a trained VGG-16 model performing on the ImageNet dataset. In order to conduct a more comprehensive analysis, we also evaluate the sensitivity of learning rates and optimization iterations. The performance is measured by Area Under The Curve (AUC) of the MoRF curve, for which {a lower value indicates a better result}. We present all the results in Tab.~\ref{tab:sensitivity}, where intensities of the color are associated with the AUC values. It can be seen that the performance of our method is not sensitive to $\lambda$.
For other hyper-parameters, the performance stays stable when
they are set in a reasonable range.

\begin{SCtable}[\sidecaptionrelwidth][h]
    \centering
    \caption{Results of sensitivity analysis for the hyper-parameters. AUC is the area under MoRF curve and lower is better. The performance is not sensitive to $\lambda$ and stays stable when choosing reasonable values for other hyper-parameters.}
    \scalebox{0.66}{
    \begin{tabular}{c|cccccccccc}
         $\gamma$ & 1 &  7.5& 14& 21& 27& 34& 40& 47& 53.5& 60 \\
        AUC & \Gray{0.4}0.40& \Gray{0.68}0.27& \Gray{0.92}0.16& \Gray{0.99}0.13 & \Gray{0.97}0.14& \Gray{0.94}0.15& \Gray{0.9}0.17& \Gray{0.88}0.18& \Gray{0.84}0.20& \Gray{0.79}0.22 \\
        \hline
        Iters & 1 &  8& 14& 21& 27& 34& 40& 47& 54& 60 \\
        AUC & \Gray{0.4}0.86& \Gray{0.64}0.29& \Gray{0.68}0.27& \Gray{0.68}0.27& \Gray{0.71}0.26& \Gray{0.75}0.24& \Gray{0.77}0.23& \Gray{0.77}0.23& \Gray{0.79}0.22& \Gray{0.77}0.23 \\
        \hline
         $\lambda$ & 0.1& 0.2& 0.3& 0.4& 0.5& 0.6& 0.7& 0.8& 0.9& 1 \\
        AUC & \Gray{0.77}0.23& \Gray{0.79}0.22& \Gray{0.79}0.22& \Gray{0.79}0.22& \Gray{0.79}0.22& \Gray{0.79}0.22& \Gray{0.81}0.21& \Gray{0.79}0.22& \Gray{0.81}0.21& \Gray{0.79}0.22 \\
        \hline
        lr & 0.01& 0.06& 0.12& 0.17& 0.23& 0.28& 0.34& 0.39& 0.45& 0.5 \\
        AUC & \Gray{0.4}0.40& \Gray{0.55}0.33& \Gray{0.64}0.29& \Gray{0.70}0.26& \Gray{0.73}0.25& \Gray{0.73}0.25& \Gray{0.82}0.21& \Gray{0.82}0.21& \Gray{0.84}0.20& \Gray{0.81}0.21 \\
        \hline
        $\alpha$ & 0.55 & 0.59& 0.64& 0.68& 0.73& 0.77& 0.82& 0.86& 0.90& 0.95 \\
        AUC & \Gray{0.4}0.47 & \Gray{0.47}0.37 & \Gray{0.62}0.30 & \Gray{0.71}0.26 & \Gray{0.77}0.23 & \Gray{0.82}0.21 & \Gray{0.86}0.19 & \Gray{0.88}0.18 & \Gray{0.88}0.18 & \Gray{0.81}0.21 \\ 
    \end{tabular}
    }
    \label{tab:sensitivity}
\end{SCtable}
\begin{SCfigure}
\centering
\includegraphics[width=0.45\textwidth]{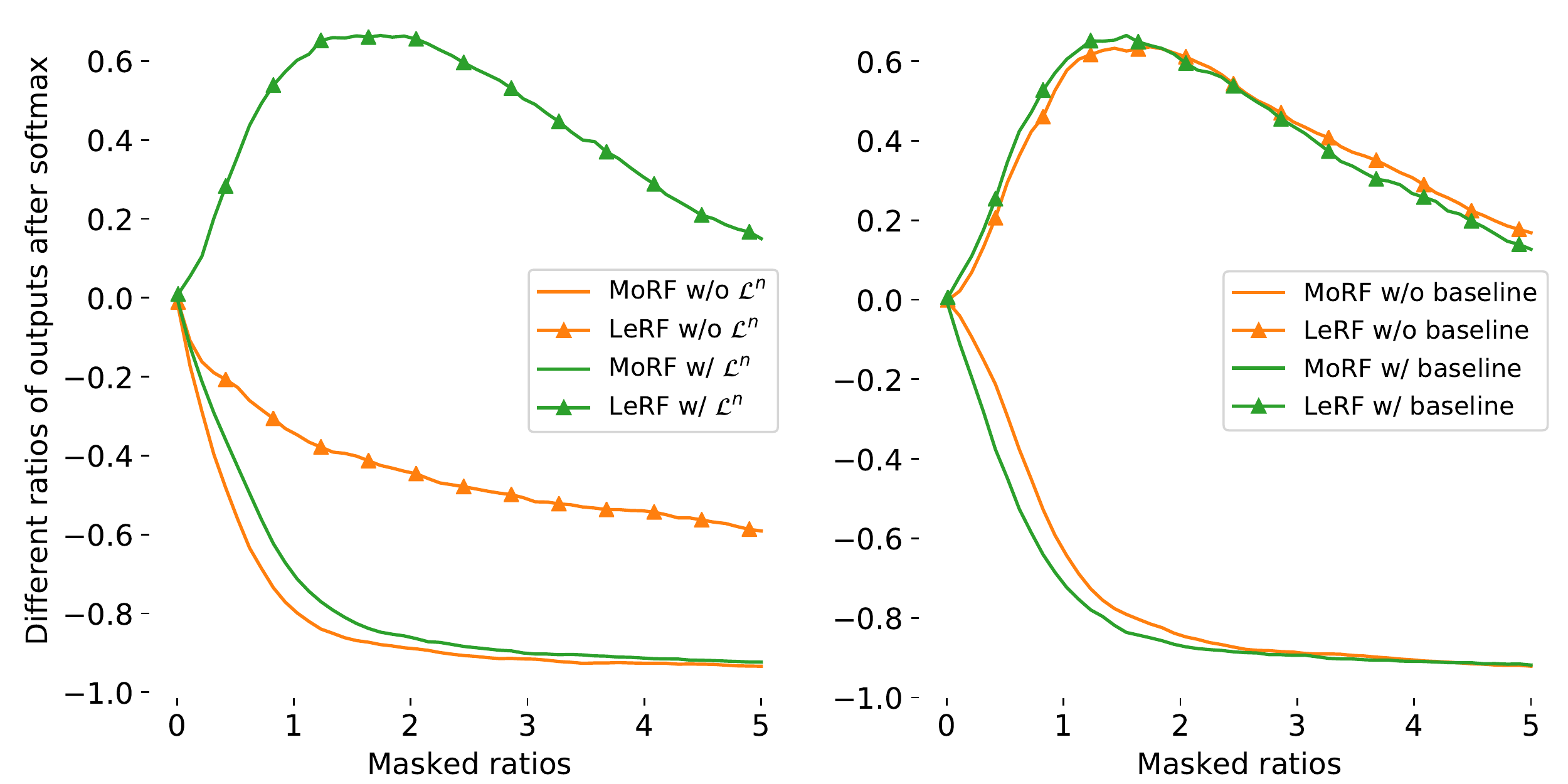} 
\caption{Ablation studies of the sign-aware loss and the reference baseline.
The sign-aware loss dramatically improves the performance on LeRF curve while the baseline reference has a little influence on the performance.}
\label{fig:ablation}
\end{SCfigure}

\paragraph{\bf Ablation Studies.}
We conduct ablation studies to analyze the effect of two terms, including the sign-aware loss and the reference baseline. The comparison results between the full model and models without one of two terms are presented in Fig.~\ref{fig:ablation}. It can be seen that the sign-aware loss improves the performance on LeRF curve by a significant margin with cost of a little affect on the performance of MoRF.
The intuition behind is that the sign-aware loss, which contains 
two branches, will generate the guided supervisions for two types of mask separately.
As for the reference baseline, it influences the performance of the proposed method 
on both LeRF and MoRF curves slightly, implying that 
the learnable plugin module is already flexible enough even without the  
cues provided by the added reference features.  
{We also tried to use the standard convolution as the plugin module but fail to
generate a meaningful attribution map. This can be partially explained by that,
the standard convolution operation does a substantial change to the gradients, 
leading to a pointless attribution map that is unrelated to the model anymore.}

\section{Conclusion}
In this paper, we propose a dedicated 
attribution-map method that
enables the propagation rules learnable 
for the non-linear layers, so as to overcome the drawbacks
of existing gradient-base methods.
The propagation rules are controlled by 
the plugin module and can be optimized
by the proposed optimization scheme under
any auto-grad framework with higher-order
differential support.
The learnable rules are adaptive, thanks to the supervision
from the model and the input themselves.
As demonstrated on several datasets and models,
our method yields state-of-the-art results
and produces cleaner and more focused attribution maps.

\paragraph{\bf Acknowledgments.}
This research is supported by the startup funding of 
Stevens Institute of Technology and Australian Research Council Projects FL-170100117, DP-180103424, IC-190100031. Xinchao Wang is the corresponding author of this paper.

\clearpage

\bibliographystyle{splncs04}
\bibliography{references}
\end{document}